%% file: sample-sigconf.tex
\documentclass[sigconf]{acmart}
\AtBeginDocument{%
  }

\setcopyright{acmlicensed}
\copyrightyear{2018}
\acmYear{2018}
\acmDOI{XXXXXXX.XXXXXXX}
\acmConference[Conference acronym 'XX]{Make sure to enter the correct
  conference title from your rights confirmation email}{June 03--05,
  2018}{Woodstock, NY}
\acmISBN{978-1-4503-XXXX-X/2018/06}

\usepackage{amsmath}
\usepackage{mathtools}
\usepackage{amsthm}
\usepackage{subcaption}
\usepackage{pifont}

\newcommand{\cmark}{\ding{51}} 
\newcommand{\xmark}{\ding{55}} 

\usepackage{multirow}

\usepackage{xspace}

\renewcommand\footnotetextcopyrightpermission[1]{}
\setcopyright{none}

\newcommand{\ours}{BindCLIP\xspace}

\begin{document}

\title{BindCLIP: A Unified Contrastive–Generative Representation Learning Framework for Virtual Screening}

\author{Anjie Qiao}
\affiliation{%
  \institution{Sun Yat-sen University}
  \city{Guangzhou}
  \country{China}
}

\author{Zhen Wang}
\affiliation{%
  \institution{Sun Yat-sen University}
  \city{Guangzhou}
  \country{China}
}

\author{Yaliang Li}
\affiliation{%
  \institution{Alibaba Group}
  \city{Hangzhou}
  \country{China}
}

\author{Jiahua Rao}
\affiliation{%
  \institution{Sun Yat-sen University}
  \city{Guangzhou}
  \country{China}
}

\author{Yuedong Yang}
\affiliation{%
  \institution{Sun Yat-sen University}
  \city{Guangzhou}
  \country{China}
}

\renewcommand{\shortauthors}{Trovato et al.}

\begin{abstract}
Virtual screening aims to efficiently identify active ligands from massive chemical libraries for a given target pocket.
Recent CLIP-style models such as DrugCLIP enable scalable virtual screening by embedding pockets and ligands into a shared space.
However, our analyses indicate that such representations can be insensitive to fine-grained binding interactions and may rely on shortcut correlations in training data, limiting their ability to rank ligands by true binding compatibility.
To address these issues, we propose \ours, a unified contrastive--generative representation learning framework for virtual screening.
\ours jointly trains pocket and ligand encoders using CLIP-style contrastive learning together with a pocket-conditioned diffusion objective for binding pose generation, so that pose-level supervision directly shapes the retrieval embedding space toward interaction-relevant features.
To further mitigate shortcut reliance, we introduce hard-negative augmentation and a ligand--ligand anchoring regularizer that prevents representation collapse.
Experiments on two public benchmarks demonstrate consistent improvements over strong baselines.
\ours achieves substantial gains on challenging out-of-distribution virtual screening and improves ligand-analogue ranking on the FEP+ benchmark.
Together, these results indicate that integrating generative, pose-level supervision with contrastive learning yields more interaction-aware embeddings and improves generalization in realistic screening settings, bringing virtual screening closer to real-world applicability.
\end{abstract}



\maketitle
\input{sections/intro}

\input{sections/related}
\input{sections/method}

\input{sections/exp}

\section{Conclusion}
In this work, we identify two key limitations of CLIP-style approaches to retrieval-based virtual screening---interaction insensitivity and shortcut reliance---which motivate injecting cues that reliably reflect binding interactions into the learned embeddings.
To this end, we introduce \ours, a unified contrastive-generative representation learning framework for retrieval-based virtual screening.
\ours addresses these limitations by training retrieval embeddings with structure-grounded supervision, improving interaction sensitivity while preserving the same efficient nearest-neighbor inference used in standard pipelines.
We also show that carefully constructed hard-negatives can further enhance the retrieval embeddings without destabilizing training.
Across benchmarks, this leads to consistently better screening performance, markedly stronger early enrichment in strict out-of-distribution tests, and more reliable ordering of closely related analogues.
Together, these results suggest that integrating structure-grounded generative supervision into contrastive retrieval learning is a practical route to interaction-aware embeddings. More broadly, our work offers the community a deployable principle: train with 3D binding evidence to suppress shortcuts, but keep inference as nearest-neighbor search—achieving robustness and early-recognition gains at the scale required for real virtual screening.

\bibliographystyle{ACM-Reference-Format}
\bibliography{sample-base}

\appendix
\input{sections/appendix}

\end{document}

%% file: sections/intro.tex
\section{Introduction}
\label{sec:intro}
Advances in synthetic chemistry have enabled much larger compound collections~\cite{grygorenko2020_synthetic1,bellmann2022_synthetic2}. Today’s chemical libraries provide access to billions of diverse molecules, creating a major opportunity for drug discovery~\cite{lyu2019_ultravr}. However, their scale makes exhaustive experimental screening infeasible in both time and cost. Thus, virtual screening is essential for leveraging this opportunity by identifying and prioritizing candidates from large chemical libraries for a given target protein~\cite{zhou2024_aivr,luttens2025_rapidvr}. 

In practice, virtual screening methods must efficiently and accurately identify a small number of active compounds from a vast majority of inactive ones~\cite{lyu2023_vr3}. 
Docking-based methods~\cite{friesner2004_glide,trott2010_autodock,alhossary2015_qvina} are widely used to estimate target-specific binding affinity but are computationally expensive at library scale.
Learning-based methods replace docking with low-cost inference and typically formulate this problem as regression~\cite{nguyen2021_graphdta,zhang2023_planet} or classification~\cite{wu2022_bridgedpi,yazdani2022_attentionsitedti,singh2023_conplex}.
Regression labels aggregated across assays are often noisy and not directly comparable due to differences in experimental protocols and incompatible measurements~\cite{landrum2024_regressiondata1,feng2024_regressiondata2}. Classification, in turn, usually requires constructing negative pairs by sampling unlabeled pairs as negatives~\cite{peng2017_classification1}.
Recently, DrugCLIP~\cite{gao2023_drugclip1,jia2026_drugclip2} reframes virtual screening as retrieval: it learns pocket and ligand representations with a CLIP-style objective~\cite{radford2021_clip}, reducing reliance on affinity values and scaling without manually curated negatives.

A key advantage of retrieval-based screening is efficiency. Given a query pocket, instead of exhaustively running inference over the entire library, DrugCLIP performs a nearest-neighbor search in the embedding space, often facilitated by a vector database~\cite{NEURIPS2019_diskann}.
This paradigm, however, critically depends on whether similarity in the embedding space aligns with binding compatibility. Since binding is driven by specific local interactions and their spatial arrangement, this requirement is nontrivial: prior work shows that CLIP-style models favor global, coarse-grained semantics and often miss fine-grained cues~\cite{rao2022_denseclip,jing2024_fineclip,monsefi2024_detailclip,xiefg2025_FGCLIP} and compositional/spatial structure~\cite{abbasi2025_scopeclip,wang2025_spatialclip}.
Moreover, in the absence of fine-grained interaction supervision (e.g., which moieties interact and why), the model may exploit dataset-specific correlations as shortcuts rather than learning genuine binding compatibility.

\begin{figure}[t]
  \centering

  \begin{subfigure}[t]{\linewidth}
    \centering
    \includegraphics[width=\linewidth]{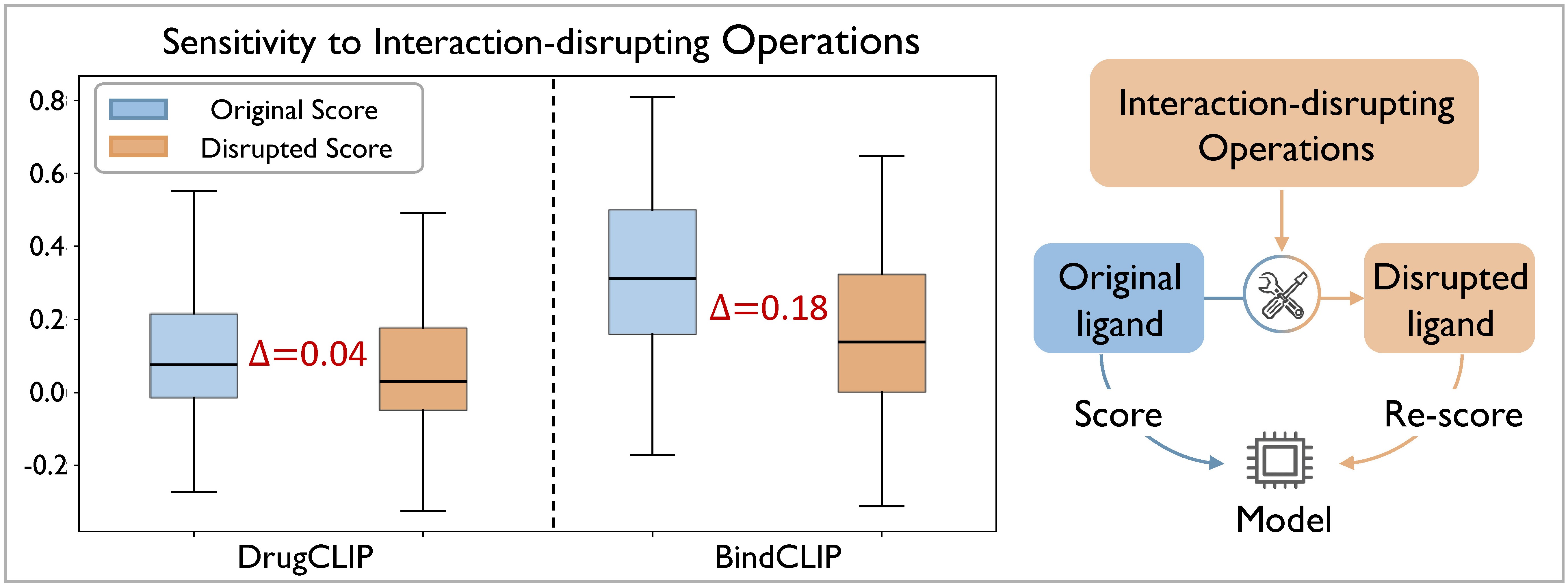}
    \caption{Whether model-assigned scores, used to rank molecules, decrease after applying interaction-disrupting operations (salt-bridge disruption, amine capping, donor removal, and hydroxyl masking). $\Delta$ denotes the mean score decrease (original to disrupted) over the evaluated samples.}

    \label{fig:motivation_a}
  \end{subfigure}


  \begin{subfigure}[t]{\linewidth}
    \centering
    \includegraphics[width=\linewidth]{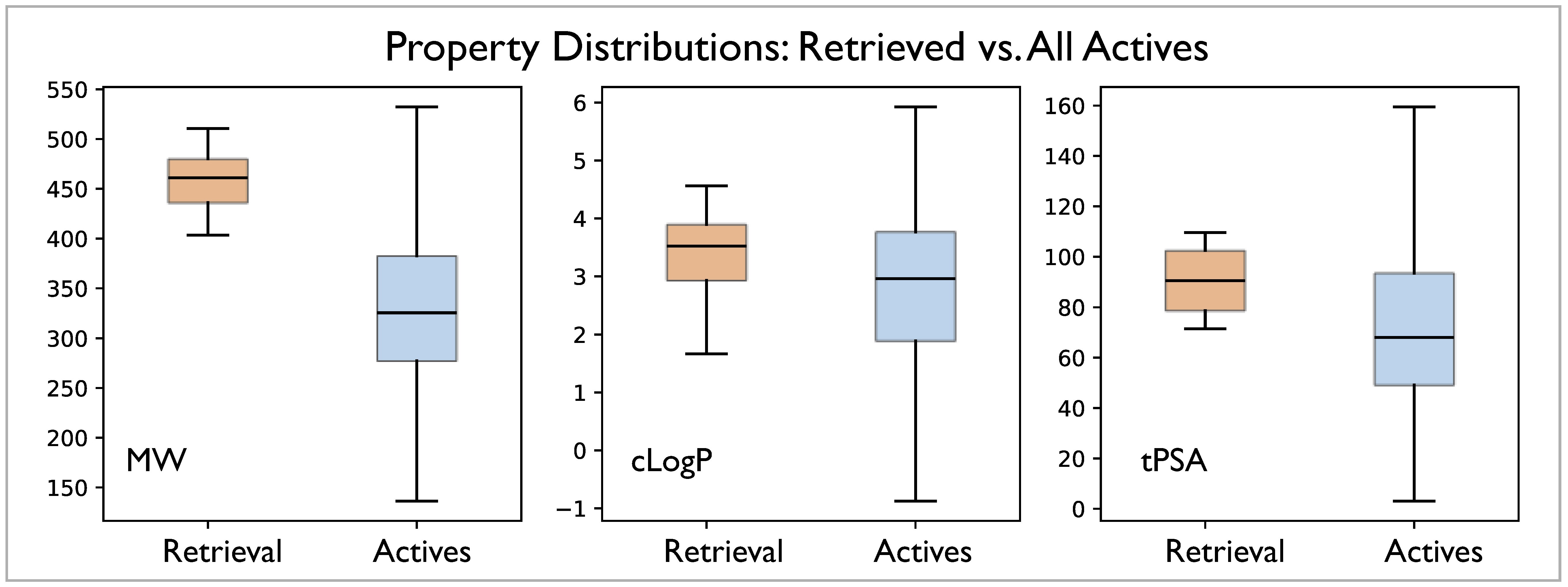}
    \caption{Distribution differences of molecular properties between retrieved candidates and ground-truth actives.}
    \label{fig:motivation_b}
  \end{subfigure}

  \caption{Illustrative probe cases from the test set.}
  \label{fig:motivation}
\end{figure}

We therefore examine two potential failure modes of CLIP-style virtual screening.
\textbf{(i) Interaction insensitivity.} We perturb active ligands with interaction-disrupting operations and evaluate whether the model assigns lower scores to the perturbed ligands.
As shown in Fig.~\ref{fig:motivation_a}, DrugCLIP exhibits limited sensitivity, suggesting that it may not adequately encode detailed interaction patterns, while our approach partially restores responsiveness to these perturbations.
\textbf{(ii) Shortcut reliance.} We further observe that DrugCLIP’s retrieved candidates occupy a more constrained region of physicochemical space than the ground-truth actives (Fig.~\ref{fig:motivation_b}), suggesting that retrieval can be dominated by coarse physicochemical similarity and may under-represent diverse actives outside this region.

These findings suggest that CLIP-style alignment alone is insufficient and motivate incorporating fine-grained binding-relevant knowledge into pocket and ligand embeddings.
Recent work on diffusion-based representation learning~\cite{fuest2024_diffsurvey,chen2025_graffe,zheng2025_novel} suggests that diffusion can serve as effective multi-scale supervision, distilling knowledge from generatively modeled variable into the conditioning representations.
In the context of virtual screening, this suggests leveraging \emph{pocket-conditioned diffusion for binding pose generation}: recent pocket-conditioned diffusion models have demonstrated strong ability to generate ligands with realistic binding poses and improved binding-related properties~\cite{guan_targetdiff,qiao2025_3d}, capturing spatially grounded pocket--ligand interactions that CLIP-style learning often misses.
Motivated by this, we propose \textbf{\ours}, a unified contrastive–generative representation learning framework for virtual screening.
\ours trains pocket and ligand encoders with a CLIP-style objective and a pocket-conditioned diffusion objective for ligand binding pose generation, enriching the embeddings with fine-grained binding signals.
Importantly, the diffusion model is used only during training; screening-time inference remains encoder embedding plus nearest-neighbor search.

To further reduce shortcut reliance and emphasize interaction-relevant distinctions, \ours incorporates hard-negative augmentation into the contrastive objective.
However, under a CLIP-style objective---with push-away-only treatment of hard negatives---they may collapse into a far-off region of the embedding space, yielding degenerate embedding geometry where hard negatives no longer provide informative gradients.
We therefore introduce an auxiliary \emph{ligand--ligand anchoring regularizer} to prevent this collapse.
We evaluate \ours on public benchmarks and more realistic and challenging settings, including stricter out-of-distribution (OOD) virtual screening and the FEP+ benchmark~\cite{ross2023_fep+}, where it consistently improves performance over strong baselines.

We summarize our contributions as follows:
\begin{itemize}
    \item We propose \ours, which leverages binding-pose generation
    as an auxiliary training objective for contrastive representation learning, yielding interaction-aware embeddings with richer pocket--ligand binding cues for virtual screening.
    \item We reinterpret pocket-conditioned diffusion as a supervision signal for representation learning, and introduce a ligand--ligand anchoring regularizer to support hard-negative augmentation.
    \item In more realistic and challenging evaluations, \ours better captures fine-grained interaction signals, improving OOD generalization and sensitivity to subtle interaction-pattern differences among ligand analogues.
\end{itemize}

\begin{figure*}[t]
  \centering
  \includegraphics[width=0.975\textwidth]{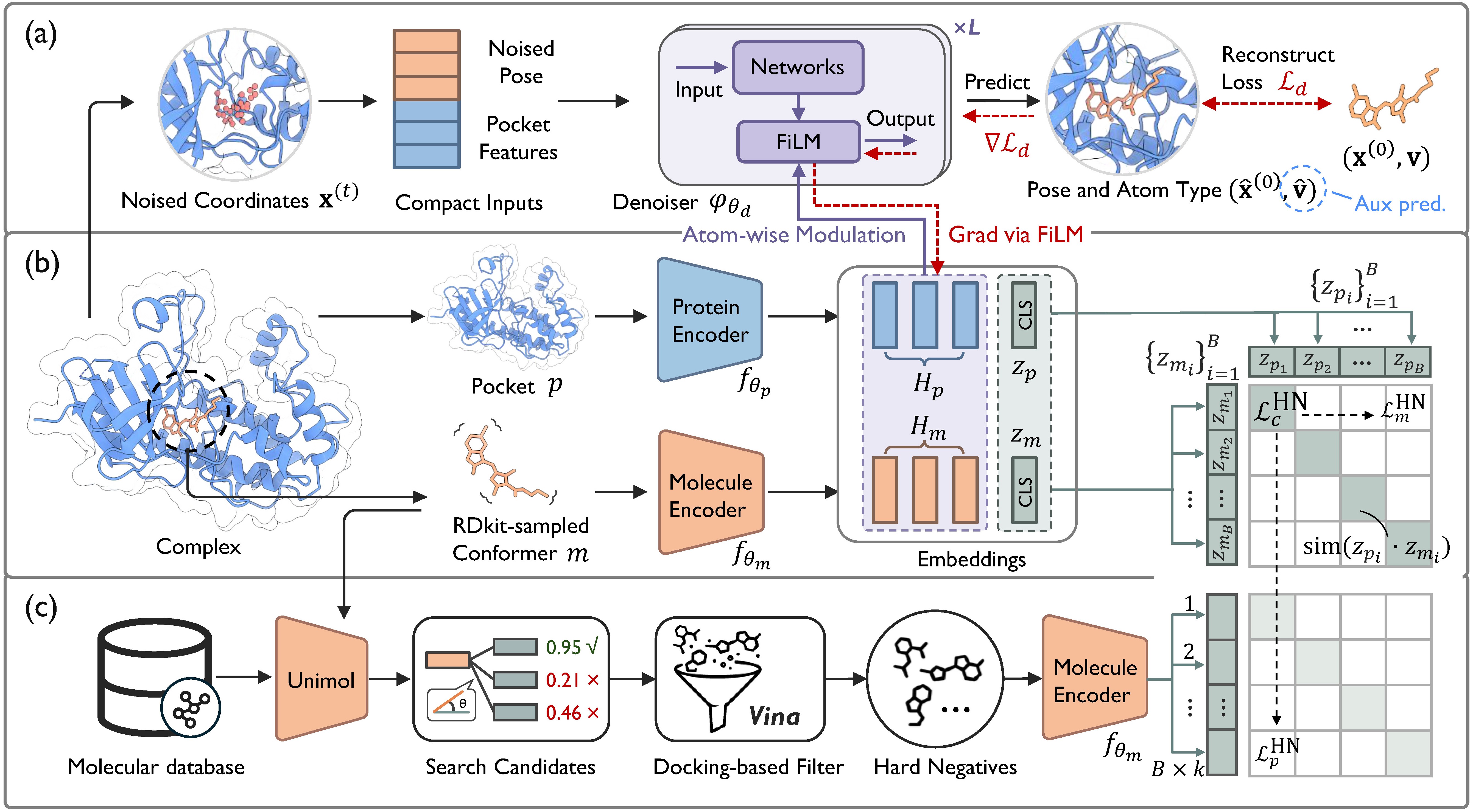}
    \caption{Overview of the \ours framework. (a) Pocket-conditioned ligand binding pose generation objective; (b) Contrastive learning objective; (c) Hard-negative augmentation workflow.}
    \label{fig:framework}
\end{figure*}

%% file: sections/related.tex
\section{Related Work}
\label{sec:related}

\noindent \textbf{Virtual Screening Methods}. Virtual screening aims to retrieve active ligands for a given target from large molecular libraries via a scoring function.
Existing methods primarily differ in how this scoring function is defined and can be broadly categorized into docking-based and learning-based approaches. 
Docking-based methods simulate the physical binding process between a ligand and a target protein to predict binding pose and energy. Classical docking approaches~\cite{friesner2004_glide,trott2010_autodock,alhossary2015_qvina} generate candidate ligand poses through sampling algorithms and evaluate binding affinity with scoring functions~\cite{verdonk2003_MonteCarlo}. This iterative procedure searches the conformational space until convergence to an optimal docking pose, making it computationally expensive and time-consuming.

Learning-based methods leverage machine learning to learn scoring functions that can be evaluated efficiently. Most prior methods treat scoring function learning as either a regression or a binary classification task. Regression approaches~\cite{ozturk2018_deepdta,zheng2019_onionnet,nguyen2021_graphdta,zhang2023_planet} predict the binding affinity of a protein-ligand complex. Classification approaches~\cite{wu2022_bridgedpi,yazdani2022_attentionsitedti,singh2023_conplex} predict whether a molecule binds to a given target. However, these supervised learning approaches exhibit limited generalization due to the scarcity of labeled data~\cite{brocidiacono2023_bigbind} and do not scale efficiently to large-scale virtual screening. Recently, DrugCLIP formulates virtual screening as an embedding-based retrieval task by learning joint protein–ligand embeddings with contrastive learning~\cite{gao2023_drugclip1,jia2026_drugclip2} and ranking candidates via embedding similarity. This formulation achieves strong performance and offers practical advantages for real-world screening scenarios, motivating subsequent studies~\cite{han2025_drughash,he2025_s2drug} that extend the CLIP-style model by adding regularization terms or incorporating protein sequence information. 
Here, we propose a unified contrastive--generative representation learning framework that combines CLIP-style learning with pocket-conditioned diffusion, yielding pocket--ligand embeddings that better reflect binding compatibility for virtual screening.

\noindent \textbf{Diffusion Models for Representation Learning}.
Recent works on diffusion models for representation learning can be broadly categorized into the following paradigms.
(1) Some approaches~\cite{xu2023_ODISE,luo2023_diffusionhyperfeatures,tang2023_DIFT} leverage intermediate representations from pre-trained diffusion models at selected timesteps and network layers as features for downstream tasks. 
(2) Other approaches~\cite{li2023_dreamteacher,yang2023_repdiffusion} distill intermediate representations learned by diffusion models into student networks, enabling efficient downstream inference without invoking diffusion models at test time.
(3) A line of work~\cite{preechakul2022_diffAE,wang2023_infodiffusion,hudson2024_soda} views diffusion models from a reconstruction or auto-encoding perspective, introducing explicit encoders or architectural constraints such that the denoising process depends on compact and decodable semantic representations.  This formulation aims to learn structured latent spaces that support both reconstruction-based generation and representation learning.
(4) Finally, some approaches~\cite{deja2023_JDM,tian2024_addp} jointly optimize generative denoising objectives and discriminative losses within a unified diffusion architecture, aiming to learn representations that are useful for both generation and recognition tasks.
Different from prior paradigms, we leverage pocket-conditioned diffusion as fine-grained geometry-aware supervision to enrich CLIP-style pocket--ligand embeddings with binding-relevant cues.

%% file: sections/method.tex
\section{Methodology}
\label{sec:method}
In this section, we first formalize retrieval-based virtual screening and summarize the CLIP-style formulation in Sec.~\ref{subsec:setup}. 
On this foundation, we develop \ours{} by introducing two key ingredients: Sec.~\ref{subsec:genobj} presents a pocket-conditioned binding pose generation objective, and Sec.~\ref{subsec:anchorreg} proposes a ligand--ligand anchoring regularizer. 
As illustrated in Fig.~\ref{fig:framework}, these components jointly steer the pocket and the ligand encoders toward interaction-relevant representations, yielding a unified framework for representation learning.

\subsection{Problem Setup and CLIP-style Approach}
\label{subsec:setup}
Let $p$ denote a query pocket and $\mathcal{M}=\{m_1,m_2,\ldots,m_n\}$ a molecular library.
Virtual screening aims to retrieve the $\text{top-}k$ molecules in $\mathcal{M}$ that are most likely to be active against $p$.
In a retrieval-based formulation, a pocket encoder $f_{\theta_p}(\cdot)$ and a molecule encoder $f_{\theta_m}(\cdot)$ map a pocket $p$ and a molecule $m$ into a shared embedding space, yielding embeddings $z_p$ and $z_m$, respectively. People often choose $\mathrm{sim}(\cdot,\cdot)$ to be a similarity that admits efficient indexing and search, such as cosine similarity (equivalently, inner product after $\ell_2\text{-normalization}$), and define the score $s(p,m)=\mathrm{sim}(z_p, z_m)$, where a larger score indicates higher binding compatibility between $p$ and $m$.
Assuming the encoders are well trained so that $s(p,m)$ reliably reflects binding compatibility, the screening result is obtained by selecting the $k$ molecules with the largest scores: $
    \mathcal{R}_k(p)=\operatorname*{argmax}_{\substack{S \subseteq \mathcal{M}\\|S| = k}}
\sum_{m \in S} s(p,m)$,
which is equivalent to ranking all $m\in \mathcal{M}$ by $s(p,m)$ and taking the top $k$. In practice, one can precompute $\{z_m\}_{m\in\mathcal{M}}$ and build an index over these embeddings; at inference time, exact or approximate nearest-neighbor (ANN) search can then efficiently return the $k$ molecules with the largest $s(p,m)$ for the query pocket $p$.

\noindent\textbf{Encoders.}
Although our framework is compatible with various encoder architectures, we follow existing CLIP-style models~\cite{gao2023_drugclip1,han2025_drughash} and instantiate both $f_{\theta_p}$ and $f_{\theta_m}$ with the pre-trained UniMol Transformer backbone~\cite{zhou2023_unimol}. Given an input pocket (resp., ligand) with atom types $v_{1:N}$ and 3D coordinates $a_{1:N}$, where the pocket coordinates are taken from the given protein structure and the ligand coordinates correspond to an \emph{unbound conformer} generated by RDKit MMFF~\cite{landrum2013_rdkit} in practice, UniMol prepends a global token with type $v_0 = \texttt{[CLS]}$ and coordinate $a_0=\frac{1}{N}\sum_{i=1}^{N}a_i$ (the center of geometry). The encoder outputs a global embedding $z$ together with atom-level embeddings $H$: 
\begin{equation}
(z, H)=f(\{(v_i,a_i)\}_{i=0}^{N}),\quad z\in\mathbb{R}^d,\; H\in\mathbb{R}^{N\times d},
\label{eq:atom_emb}
\end{equation}
where $z$ corresponds to the \texttt{[CLS]} token and $H$ stacks the embeddings of the $N$ atoms.
UniMol treats the \texttt{[CLS]} token as a special ``atom'', which couples the global embedding $z$ with the atom-level embeddings $H$ through shared model parameters and attention mechanism; therefore, optimizing one will generally influence the other.
For brevity, we denote $(z_p, H_p)=f_{\theta_p}(p)$ and $(z_m, H_m)=f_{\theta_m}(m)$.
In addition to the global embeddings used for contrastive training below, \ours{} later leverages the atom-level embeddings $(H_p, H_m)$ to condition a diffusion denoiser for auxiliary supervision (Sec.~\ref{subsec:genobj}).

\noindent\textbf{Contrastive Objective.}
To train such encoders so that the resulting score $s(p,m)$ can reliably reflect binding compatibility, CLIP-style contrastive learning with the InfoNCE objective~\cite{radford2021_clip} is commonly adopted in the literature. In many pocket-ligand datasets, supervision is primarily provided as observed binding (positive) pairs, while explicit non-binding (negative) labels are unavailable.
Concretely, given a mini-batch of $B$ positive pairs $\{(p_i,m_i)\}_{i=1}^{B}$, embeddings are obtained as $z_{p_i}=f_{\theta_p}(p_i)$ and $z_{m_i}=f_{\theta_m}(m_i)$. Each matched pair $(p_i,m_i)$ serves as a positive example, whereas mismatched pairs $(p_i,m_j)$ and $(p_j,m_i)$ for $j\neq i$ are treated as in-batch negatives (i.e., unobserved pairs assumed to be negative). The similarity is typically instantiated as cosine similarity,
\begin{equation}
\label{eq:similarity}
    s(p_i,m_j) = \frac{(z_{p_i})^{\top} \cdot z_{m_j}}{\lVert z_{p_i}\rVert\,\lVert z_{m_j}\rVert}.
\end{equation}
Accordingly, the symmetric InfoNCE losses from the pocket and molecule perspectives are written as:
\begin{equation}
\begin{aligned}
\label{eq:contrastive1}
\mathcal{L}_{p}(\theta_p, \theta_m)
= -\frac{1}{B}\sum_{i=1}^{B}
\log \frac{\exp(s(p_i,m_i)/\tau)}{\sum_{j=1}^{B}\exp(s(p_i,m_j)/\tau)}, \\
\mathcal{L}_{m}(\theta_p, \theta_m)
= -\frac{1}{B}\sum_{i=1}^{B}
\log \frac{\exp(s(p_i,m_i)/\tau)}{\sum_{j=1}^{B}\exp(s(p_j,m_i)/\tau)},
\end{aligned}
\end{equation}
where $\tau$ is a temperature, and the overall contrastive objective is:
\begin{align}
\label{eq:contrastive2}
\mathcal{L}_{c}(\theta_p,\theta_m) = \frac{1}{2}\big(\mathcal{L}_{p}(\theta_p, \theta_m)+\mathcal{L}_{m}(\theta_p, \theta_m)\big).
\end{align}

\subsection{Binding Pose Generation-guided Representation Learning}
\label{subsec:genobj}
The contrastive objective in Eq.~\ref{eq:contrastive2} aligns the \emph{global} pocket and ligand embeddings $(z_p, z_m)$ for observed binding pairs, but it does not explicitly supervise the \emph{atom-level embeddings} needed to capture fine-grained pocket--ligand interactions. 
To inject such interaction-relevant signals into the encoders, we introduce a binding pose generation task and instantiate it with a diffusion model that learns a conditional distribution over the binding pose given a pocket, the ligand atom types, and a non-binding conformer of the ligand.
Concretely, the diffusion denoiser is trained to reconstruct a noised ligand pose under these conditions, where the conditioning leverages the pocket atom-level embeddings produced by $f_{\theta_p}$ and the ligand atom-level embeddings produced by $f_{\theta_m}$. 
This generative objective is used only during training to shape the encoders; inference remains unchanged and does not involve the denoiser.

\noindent\textbf{Forward process.}
We represent a ligand binding pose as $\mathbf{x}^{(0)} \in \mathbb{R}^{N_m \times 3}$, corresponding to the coordinates of the $N_m$ ligand atoms observed in the pocket--ligand complex.
For diffusion step $t\in\{1,\dots,T\}$, we corrupt the continuous coordinates with Gaussian noise, following a fixed noise schedule~\cite{hoogeboom2021_scheme}.
This defines a forward (diffusion) process with a closed-form marginal transition $q(\mathbf{x}^{(t)}\mid \mathbf{x}^{(0)})$.
Details are provided in Appendix~\ref{appendix:diffusion}.

\noindent\textbf{Reverse process with atom-wise conditioning.}
In the reverse (denoising) process, the clean binding pose $\mathbf{x}^{(0)}$ is generated from the initial noise $\mathbf{x}^{(T)}$ via a sequence of reverse transitions
$p(\mathbf{x}^{(t-1)} \mid \mathbf{x}^{(t)},p,m)$ for $t=T,\dots,1$.
We parameterize these reverse transitions with an SE(3)-equivariant
denoiser $\phi_{\theta_d}$ that predicts the clean binding pose
$\mathbf{x}^{(0)}$ conditioned on the pocket $p$, timestep $t$, and atom-level embeddings $(H_p, H_m)$ produced by the pocket and ligand encoders: $\hat{\mathbf{x}}^{(0)}=\phi_{\theta_d}(\mathbf{x}^{(t)},p,t, H_p, H_m)$, so that the encoders participate in the denoising process and thus can receive gradients from the generative objective.

Specifically, the denoiser $\phi_{\theta_d}$ consists of input/output modules $\psi_{\mathrm{in}}$ and $\psi_{\mathrm{out}}$, a stack of $L$ SE(3)-equivariant blocks $\{F_\ell\}_{\ell=1}^L$, and layer-specific MLPs $\{g_\ell\}_{\ell=1}^L$ used for Feature-wise Linear Modulation (FiLM)~\cite{perez2018_film}.
The denoiser maintains an atom-wise hidden state $h_{\ell}\in\mathbb{R}^{(N_p+N_m)\times d_h}$ in each layer.
Particularly, we initialize $h_{0}^{(t)}$ from inputs: $h_{0} =\psi_{\mathrm{in}}(\mathbf{x}^{(t)}, p, t)\in\mathbb{R}^{(N_p+N_m)\times d_h}$.
For $\ell=1,\ldots,L$, we apply the $\ell$-th SE(3)-equivariant block: $\tilde h_{\ell} = F_\ell\!\left(h_{\ell-1}\right)$.
Then, the atom-level embeddings $H_p$ and $H_m$ are concatenated to form $[H_p; H_m] \in \mathbb{R}^{(N_p+N_m)\times d}$. A layer-specific MLP $g_\ell : \mathbb{R}^d \rightarrow \mathbb{R}^{2d_h}$ maps each atom embedding to FiLM parameters, which are used to perform an \emph{atom-wise FiLM modulation} on the hidden states:
\begin{equation}
\begin{aligned}
&[\gamma_{\ell,i}, \beta_{\ell,i}] = g_\ell\big([H_p; H_m]_i\big),
\qquad \gamma_{\ell,i}, \beta_{\ell,i} \in \mathbb{R}^{d_h},
\\
&h_{\ell,i} = \gamma_{\ell,i} \odot \tilde{h}_{\ell,i} + \beta_{\ell,i},
\qquad i \in \{1,\ldots, (N_p + N_m)\}.
\end{aligned}
\end{equation}
Thus, each atom receives a modulation signal derived from the corresponding pocket or ligand atom embedding at every layer.

After $L$ blocks, we predict binding ligand variables from the ligand portion of the hidden state.
Let $\mathcal{I}_m=\{N_p+1,\ldots,N_p+N_m\}$ denote the ligand atom indices, and define
$h_{L,m} = h_{L}[\mathcal{I}_m]\in\mathbb{R}^{N_m\times d_h}.$
We then predict the binding pose:
$
\hat{\mathbf{x}}^{(0)} = \psi_{\mathrm{out}}\!\left(h_{L,m}\right), 
$
where $\hat{\mathbf{x}}^{(0)}\in\mathbb{R}^{N_m\times 3}$.
Moreover, we observe that the contrastive objective (Eq.~\ref{eq:contrastive1}) tends to make ligand atom-level embeddings fail to capture atom type-related information (see Appendix~\ref{appendix:atomlevel}).
To counteract this side effect, we introduce an additional prediction head $\psi_{\mathrm{type}}$ that predicts ligand atom types alongside binding pose prediction:
$
\hat{\mathbf{v}} = \psi_{\mathrm{type}}\!\left(h_{L,m}^{(t)}\right), 
$
where $\hat{\mathbf{v}}\in\mathbb{R}^{N_m\times K}$ and $K$ is the number of atom types.


\noindent\textbf{Generative objective.}
The diffusion loss consists of a coordinate reconstruction term and a
cross-entropy term for atom-type prediction, weighted by $\lambda_{\mathrm{type}}$:
\begin{equation}
\begin{aligned}
\mathcal{L}_{\mathrm{d}}(\theta_{d},\theta_{p},\theta_{m})
&=
\left\| \mathbf{x}^{(0)} - \hat{\mathbf{x}}^{(0)} \right\|_2^2
\;+\;
\lambda_{\mathrm{type}}
\mathrm{CE}\!\left( \mathbf{v}, \hat{\mathbf{v}} \right).
\end{aligned}
\label{eq:genobj}
\end{equation}

\noindent\textbf{Interpretation.}
Following the analysis of conditional diffusion-guided representation learning~\cite{chen2025_graffe,zheng2025_novel}, the diffusion objective can be roughly interpreted as maximizing conditional mutual information between the clean target and the conditioning representation, given the noisy target.
In the case of \ours, when the denoiser predicts $\hat{\mathbf{x}}^{(0)}$ from
$\mathbf{x}^{(t)}$ conditioned on $[H_p; H_m]$, minimizing the
reconstruction objective (Eq.~\ref{eq:genobj}) encourages an increase of $I(\mathbf{x}^{(0)};\,[H_p;H_m]\mid \mathbf{x}^{(t)})$.
Therefore, the conditioning atom-level embeddings are incentivized to encode information that is predictive of the clean binding pose.
Because $\mathbf{x}^{(0)}$ is determined by pocket--ligand compatibility, such information corresponds to interaction-relevant supervision.

\subsection{Hard-Negative Augmentation with Ligand-Ligand Anchoring Regularizer}
\label{subsec:anchorreg}
While the diffusion objective provides fine-grained, interaction-relevant supervision, the contrastive objective operates on global embeddings and may therefore be dominated by coarse \emph{and potentially spurious} discriminative cues (i.e., shortcut signals), such as overall chemical similarity (see Fig.~\ref{fig:motivation_b}).
To sharpen the contrastive signal, we augment training with \emph{hard-negatives}--ligands that closely resemble the positive ligand--thereby explicitly enforcing discrimination within a narrow region of chemical space where shortcut cues are less informative.
In the following, we address three practical questions: (i) how to mine hard-negatives, (ii) how to incorporate them into the contrastive objective, and (iii) how to mitigate the failure modes that hard-negatives may introduce, such as representation collapse.

\noindent\textbf{Hard-negative mining.}
For each positive pair $(p, m)$, we aim to mine $k$ hard negatives $\{\tilde{m_i}\}_{i=1}^k$ from a large drug-like molecular library.
However, two key considerations arise.
First, as shown in Fig.~\ref{fig:motivation_b}, it is unclear which physicochemical properties the encoders will rely on, and to what extent, making it difficult to define a property-based matching criterion.
Second, property-based matching methods, such as per-property filtering against molecular libraries, are computationally expensive at scale.

To address both, we cast hard-negative mining as an embedding-based retrieval task using the UniMol molecular encoder. 
UniMol is well suited for this purpose, as it is pretrained to capture both physicochemical and 3D structural signals in its representations, allowing similarity to be assessed implicitly via proximity in the molecular embedding space.
Specifically, we encode each molecule in the library and retrieve candidate molecules for a query ligand via nearest-neighbor search in the embedding space.

Moreover, to reduce the risk of false negatives, we dock candidates to the pocket $p$ using AutoDock Vina~\cite{trott2010_autodock}, and filter out candidates with docking scores better than that of the (positive) query ligand.
Among the remaining candidates, we select the $k$ nearest neighbors as hard negatives $\{\tilde{m}_i\}_{i=1}^k$.
This yields a set of challenging yet less noisy hard negatives per positive ligand.

\begin{figure}[t]
  \centering
  \includegraphics[width=\linewidth]{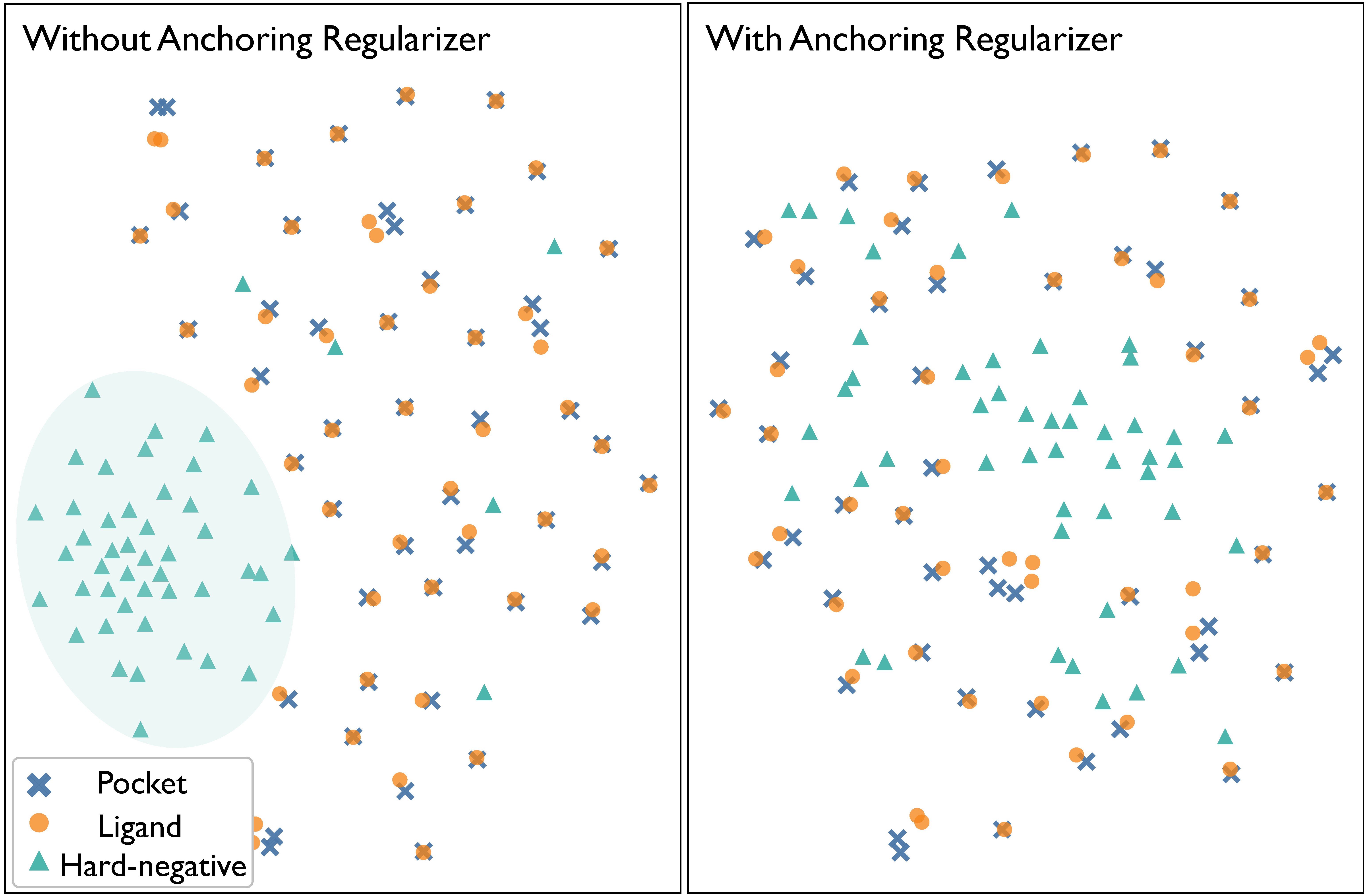}
    \caption{Visualization of pocket, ligand, and hard-negative embeddings with/without the anchoring regularizer (t-SNE).}
    \label{fig:hardneg}
\end{figure}

\noindent\textbf{Hard-negative augmented contrastive objective.}
Given a mini-batch of $B$ positive pairs $\{(p_i,m_i)\}_{i=1}^{B}$ and mined hard negatives $\{\tilde{m}_i\}_{i=1}^{B\times k}$, we treat $\tilde{m}_i$ as additional negatives for the pocket-side loss $\mathcal{L}_p$ and extend the softmax denominator:
\begin{equation}
\begin{aligned}
\label{eq:hardneg_pocket}
\mathcal{L}_{p}^{\mathrm{HN}}=
\frac{-1}{B}\sum_{i=1}^{B}
\log
\frac{\exp(s(p_i,m_i)/\tau)}
{\sum_{j=1}^{B}\exp(s(p_i,m_j)/\tau)
+
\sum_{j=1}^{B\times k}\exp(s(p_i,\tilde{m}_j)/\tau)}.
\end{aligned}
\end{equation}
We keep the ligand-side loss $\mathcal{L}_m$ unchanged (Eq.~\ref{eq:contrastive1}) since hard-negatives are not positively matched with pockets in the mini-batch:
$\mathcal{L}_{m}^{\mathrm{HN}}=\mathcal{L}_{m}.$
Thus, the hard-negative augmented contrastive objective is
\begin{equation}
\mathcal{L}_{\mathrm{c}}^{\mathrm{HN}}(\theta_p, \theta_m)=\tfrac{1}{2}\left(\mathcal{L}_{p}^{\mathrm{HN}}(\theta_p, \theta_m)+\mathcal{L}_{m}^{\mathrm{HN}}(\theta_p, \theta_m)\right).
\label{eq:hardcl}
\end{equation}

\noindent\textbf{Ligand--ligand anchoring regularizer.}
Because hard-negatives are treated as negatives for all pockets, they consistently receive repulsive gradients under the contrastive objective, causing them to collapse toward a distant region of the embedding space (Fig.~\ref{fig:hardneg}).

To alleviate this issue, we introduce a ligand-ligand anchoring regularizer that encourages each augmented hard negative ligand to remain within a comparable neighborhood of its corresponding positive ligand.
In a mini-batch of $B$ positive pairs, let $\{\tilde{m}_{i,\ell}\}_{i=1,\ell=1}^{B,k}$ denote the $k$ hard negatives for each ligand $m_i$, and
$z_{\tilde{m}_{i,\ell}} = f_{\theta_m}(\tilde{m}_{i,\ell})$ their corresponding embeddings.
We denote the similarity between each ligand and its corresponding hard negative as
$s_{i}^{\mathrm{hard}}
=
\max_{\ell\in\{1,\dots,k\}}
\mathrm{sim}\!\left(z_{m_i}, z_{\tilde{m}_{i,\ell}}\right),
$
and an in-batch reference similarity $\bar{s}_i = \frac{1}{B-1}\sum_{j\neq i}\mathrm{sim}(z_{m_i},z_{m_j})$, which reflects the expected similarity between $m_i$ and a randomly sampled molecule.
We then impose a hinge-style regularizer:
\begin{equation}
\mathcal{L}_{\mathrm{a}}(\theta_m)
=
\sum_{i=1}^{B}
\max\bigl(0,\;\mathrm{sg}(\bar{s}_i) - s_i^{\mathrm{hard}} + \delta\bigr),
\label{eq:anchor}
\end{equation}
with margin $\delta\ge 0$ and $\mathrm{sg}(\cdot)$ denotes the stop-gradient operator.

Intuitively, the regularizer prevents hard-negatives from being pushed
too far away from their corresponding query ligands, encouraging hard-negative
embeddings to remain closer than a random one while still being separated
from the positives, thereby reducing the tendency toward shortcut solutions
in contrastive learning.

\subsection{Training and Inference}
\label{subsec:summary}
During training, we sample a mini-batch of $B$ pocket--ligand pairs $\{(p_i,m_i)\}_{i=1}^{B}$ together with $k$ mined hard-negative ligands $\{\tilde m_{i,\ell}\}_{\ell=1}^{k}$ for each $m_i$.
For each pocket--ligand complex, we have access to the pocket structure and the ligand binding pose from the resolved complex.
In addition, we sample an \emph{unbound} conformer for each ligand $m_i$ and each hard-negative $\tilde m_{i,\ell}$ via RDKit MMFF.
As detailed in the explanation of Eq.~\ref{eq:atom_emb}, the protein encoder $f_{\theta_{p}}$ takes the atom coordinates of the pocket as input, while the molecule encoder $f_{\theta_m}$ consumes the unbound version.

Then, we encode $p_i$, $m_i$, and $\tilde m_{i,\ell}$, obtaining
$(z_{p_i},H_{p_i})=f_{\theta_p}(p_i)$,
$(z_{m_i},H_{m_i})=f_{\theta_m}(m_i)$,
and $(z_{\tilde m_{i,\ell}},H_{\tilde m_{i,\ell}})=f_{\theta_m}(\tilde m_{i,\ell})$.
The resulting global embeddings are used to compute the hard-negative augmented contrastive loss $\mathcal{L}_{\mathrm{c}}^{\mathrm{HN}}$ (Eq.~\ref{eq:hardcl}) and regularizer $\mathcal{L}_{\mathrm{a}}$ (Eq.~\ref{eq:anchor}), while the atom-level embeddings $(H_{p_i},H_{m_i})$ provide conditioning for the denoiser.
To train the denoiser, we set $\mathbf{x}_i^{(0)}$ to be the observed binding pose, sample $t\sim\mathcal{U}(\{1,\ldots,T\})$, and sample $\mathbf{x}_i^{(t)}$ according to the forward marginal transition.
With the prediction $(\hat{\mathbf{x}}_{i}^{(0)}, \hat{\mathbf{v}}_{i})=\phi_{\theta_d}(\mathbf{x}_{i}^{(t)}, p_i, t, H_{p_i}, H_{m_i})$, we are allowed to compute the diffusion loss $\mathcal{L}_{\mathrm{d}}$ (Eq.~\ref{eq:genobj}).
All components are optimized jointly under the unified objective:
\begin{equation}
\mathcal{L}(\theta_p,\theta_m,\theta_d)
=
\mathcal{L}_{\mathrm{c}}^{\mathrm{HN}}(\theta_p,\theta_m)
+
\lambda_{\mathrm{d}}\mathcal{L}_{\mathrm{d}}(\theta_d,\theta_p,\theta_m)
+
\lambda_{\mathrm{a}}\mathcal{L}_{\mathrm{a}}(\theta_m).
\end{equation}
We train with the Adam~\cite{2015kingma_adam} optimizer. Hyper-parameters such as $\tau$, $\lambda_{\mathrm{d}}$, $\lambda_{\mathrm{a}}$ (and $\lambda_{\mathrm{type}}$) are selected based on validation performance; additional implementation details are provided in Appendix~\ref{appendix:train}.

At inference, we perform virtual screening using only the trained pocket and ligand encoders, $f_{\theta_p}$ and $f_{\theta_m}$, according to Eq.~\ref{eq:similarity}, incurring no additional inference cost.

%% file: sections/exp.tex
\section{Experiments}
\label{sec:exp}
We conduct experiments to answer three key questions:
(RQ1) How does \ours perform overall, particularly in challenging out-of-distribution virtual screening settings and ligand-analogue ranking tasks? 
(RQ2) How much does each module—diffusion supervision, hard-negative augmentation, and the anchoring regularizer—contribute to performance?
(RQ3) How can we verify that the learned embeddings truly capture richer interaction-relevant signals? We report an efficiency analysis in Appendix~\ref{appendix:efficiency}.
 
\subsection{Experimental Setup}
\noindent \textbf{Datasets.}
For a fair comparison, we adopt the same training and validation datasets as DrugCLIP. We train on PDBBind 2019~\cite{wang2005_pdbbind} augmented with HomoAug~\cite{gao2023_drugclip1}, which expands the training data to 66,164 protein-ligand complexes by incorporating homologous proteins. We use CASF-2016~\cite{su2018_casf2016} as the validation set for checkpoint and hyperparameter selection. In addition, we randomly sample 1 million unique molecules from the ZINC~\cite{irwin2020_zinc20} database to construct an external library for hard-negative augmentation.

To evaluate the virtual screening performance, we use two widely adopted benchmarks: DUD-E~\cite{mysinger2012_dude} and LIT-PCBA~\cite{tran2020_pcba}. DUD-E includes 102 targets and 22,886 active ligands, with 50 property-matched, topologically dissimilar decoys per active ligand treated as presumed inactives. LIT-PCBA is a more challenging benchmark designed to mitigate chemical biases reported in classical datasets such as DUD-E. Built from confirmatory dose–response PubChem BioAssay data, it defines actives and inactives based on experimental evidence and filters out assay artifacts, frequent hitters, and false positives, resulting in 15 targets, with 7,844 confirmed actives and 407,381 confirmed inactives.

To evaluate out-of-distribution (OOD) generalization in experimentally realistic settings, we follow Boltz-2~\cite{passaro2025_boltz2} to construct an OOD benchmark derived from MF-PCBA~\cite{buterez2023_mfpcba}.
Although MF-PCBA contains 60 biochemical assays, only 11 have associated high-quality protein–ligand complex structures available from UniProt~\cite{uniprot2015_uniprot}; we enforce a strict OOD split by excluding proteins with $>30\%$ sequence identity to any protein in our training set, resulting in one assay for evaluation.
The resulting benchmark comprises 676 active and 96,278 inactive compounds.

Accurately ranking analogues within a chemical series is a critical and challenging task as it requires distinguishing subtle differences in binding patterns among closely related compounds.
To evaluate this capability, we follow Boltz-2 and adopt the FEP+ benchmark~\cite{ross2023_fep+}, using the 4-target FEP subset (CDK2, TYK2, JNK1, and P38)~\cite{hahn2022_fep}.

\begin{table}[t]
\caption{Results on LIT-PCBA. Best results are shown in bold.}
\label{table:PCBA}
\begin{center}
\small
\begin{tabular}{cccccc}
\toprule
         & AUROC & BEDROC & $\text{EF}^{0.5\%}$ & $\text{EF}^{1\%}$ & $\text{EF}^{5\%}$ \\ \midrule
Surflex  & 51.47 & -      & -       & 2.50  & -     \\
Glide-SP & 53.15 & 4.00   & 3.17    & 3.41  & 2.01  \\ \midrule
Planet   & 57.31 & -      & 4.64    & 3.87  & 2.43  \\
Gnina    & \textbf{60.93} & 5.40   & -       & 4.63  & -     \\
DeepDTA  & 56.27 & 2.53   & -       & 1.47  & -     \\
BigBind  & 60.80 & -      & -       & 3.82  & -     \\
DrugHash & $54.34_{ \pm 0.90}$  &   $6.49_{ \pm 0.40}$   &  $7.86_{ \pm 0.56}$      &   
$5.35_{\pm 0.28}$    &    $2.32_{\pm 0.18} $  \\
DrugCLIP & $55.45_{\pm 1.31} $  &   $6.41_{\pm 0.87} $   &  $8.24_{\pm 1.52} $      &   
$5.21_{\pm 1.11} $    &    $2.14_{\pm 0.35} $  
\\ \midrule
   \ours      & $59.15_{\scriptscriptstyle\pm 1.26}$  
& $\mathbf{7.88}_{\scriptscriptstyle\pm 0.25}$  
& $\mathbf{9.84}_{\scriptscriptstyle\pm 0.65}$  
& $\mathbf{6.26}_{\scriptscriptstyle\pm 0.35}$  
& $\mathbf{2.90}_{\scriptscriptstyle\pm 0.18}$  \\\bottomrule
\end{tabular}
\end{center}
\end{table}


\begin{table}[t]
\caption{Results on DUD-E. Best results are shown in bold.}
\vspace{-6pt}
\label{table:DUDE}
\begin{center}
\small
\begin{tabular}{cccccc}
\toprule
         & AUROC & BEDROC & $\text{EF}^{0.5\%}$ & $\text{EF}^{1\%}$ & $\text{EF}^{5\%}$ \\ \midrule
Vina  & 71.60 & -      & 9.13       & 7.32  & 4.44     \\
Glide-SP & 76.70 & 40.70   & 19.39    & 16.18  & 7.23  \\ \midrule
RFscore   & 65.21 & 12.41     & 4.90    & 4.52  & 2.98  \\
Pafnucy    & 63.11 & 16.50   & 4.24       & 3.86  & 3.76     \\
OnionNet  & 59.71 & 8.62   & 2.84      & 2.84  & 2.20     \\
Planet  & 71.60 & -      & 10.23       & 8.83  & 5.40     \\
DrugHash & $80.05_{\scriptscriptstyle\pm 0.48}$ & $47.22_{\scriptscriptstyle\pm 0.96}$ & $37.28_{\scriptscriptstyle\pm 1.61}$ & $29.57_{\scriptscriptstyle\pm 0.64}$ & $9.37_{\scriptscriptstyle\pm 0.16}$ \\
DrugCLIP & $79.29_{\scriptscriptstyle\pm 1.59}$ & $47.53_{\scriptscriptstyle\pm 2.73}$ & $37.90_{\scriptscriptstyle\pm 1.67}$ & $30.52_{\scriptscriptstyle\pm 1.80}$ & $10.08_{\scriptscriptstyle\pm 0.59}$ \\\midrule
\ours & $\mathbf{80.14}_{\scriptscriptstyle\pm 1.56}$ & $\mathbf{49.73}_{\scriptscriptstyle\pm 1.75}$ & $\mathbf{39.82}_{\scriptscriptstyle\pm 0.70}$ & $\mathbf{32.16}_{\scriptscriptstyle\pm 1.03}$ & $\mathbf{10.44}_{\scriptscriptstyle\pm 0.50}$ \\
 \bottomrule
\end{tabular}
\end{center}
\end{table}

\noindent\textbf{Baselines.}
On DUD-E, we compare against eight baselines, which include two docking-based methods (AutoDock Vina~\cite{trott2010_autodock} and Glide-SP~\cite{halgren2004_glidesp}), and six learning-based methods (RFscore~\cite{ballester2010_rfscore}, Pafnucy~\cite{stepniewska2017_pafnucy}, OnionNet~\cite{zheng2019_onionnet}, Planet~\cite{zhang2023_planet}, DrugCLIP~\cite{gao2023_drugclip1} and DrugHash~\cite{han2025_drughash}). On LIT-PCBA, we also compare against eight baselines, which include two docking-based methods (Surflex~\cite{spitzer2012_surflex} and Glide-SP), and six learning-based methods (Planet, Gnina~\cite{mcnutt2021_gnina}, DeepDTA~\cite{ozturk2018_deepdta}, BigBind~\cite{brocidiacono2023_bigbind}, DrugCLIP and DrugHash).
Specifically, DrugCLIP is run using its official codebase, whereas DrugHash is reproduced from its original paper since no public implementation is available.

For the OOD and FEP+ benchmarks, we compare against DrugCLIP, the closest CLIP-style retriever we build upon and a strong, reproducible baseline on public benchmarks.

\noindent \textbf{Evaluation protocol.}
We train and evaluate DrugCLIP, DrugHash, and \ours under the same experimental setting using five random seeds, reporting results as mean $\pm$ standard deviation.
We report AUROC, BEDROC, and Enrichment Factor (EF). BEDROC is well suited to virtual screening because it emphasizes early recognition by exponentially weighting the top-ranked compounds. We use the common $\text{BEDROC}_{85}$ setting, in which the top 2\% of ranked candidates contribute 80\% of the BEDROC score~\cite{truchon2007_bedroc}. EF quantifies early enrichment by comparing the fraction of actives in the top $x\%$ of the ranked list to that in the full database: $\text{EF}^{x\%}=\frac{\text{Hits}^{x\%}/\text{N}^{x\%}}{\text{Hits}^{\text{total}}/\text{N}^{\text{total}}}$, where $\text{Hits}^{x\%}$ and $\text{N}^{x\%}$ denote the number of actives and total molecules in top $x\%$, and $\text{Hits}^{\text{total}}$ and $\text{N}^{\text{total}}$ are the corresponding counts in the full database. We report the $\text{EF}^{0.5\%}$, $\text{EF}^{1\%}$, and $\text{EF}^{5\%}$.

In the OOD setting, we additionally report $\text{Hits@}K$, which measures the number of active compounds retrieved among the top-$K$ candidates.

In the FEP benchmark, we evaluate pairwise ranking accuracy on $\Delta\Delta G$ edges between ligand pairs. 
We further report Kendall tau correlation between the model-induced ranking and the ordering implied by $\Delta\Delta G$ relations, to measure free-energy ordering consistency.

\subsection{(RQ1) Results and Analysis}
\label{exp:results}
\noindent\textbf{Performance on LIT-PCBA and DUD-E Benchmarks.}
Results are reported in Table~\ref{table:PCBA} and Table~\ref{table:DUDE}.
Overall, \ours consistently outperforms DrugCLIP across all evaluation metrics.
Notably, the largest improvement is observed on the more challenging LIT-PCBA benchmark,
where \ours achieves a relative gain of 6.7\% in AUROC.
We also observe a 24.5\% average relative improvement in early enrichment metrics (BEDROC and EF).

The relatively smaller gains on DUD-E are expected.
Prior studies~\cite{chen2019_dudebias} have reported that DUD-E contains chemical biases,
which may allow models to achieve strong performance by exploiting shortcut signals.
In contrast, both our generative objective and hard-negative augmentation are designed to discourage such shortcuts,
forcing the model to capture finer-grained cues for discrimination. Thus, we next examine the OOD performance of \ours in realistic experimental cases.

\begin{figure}[t]
  \centering
  \includegraphics[width=\linewidth]{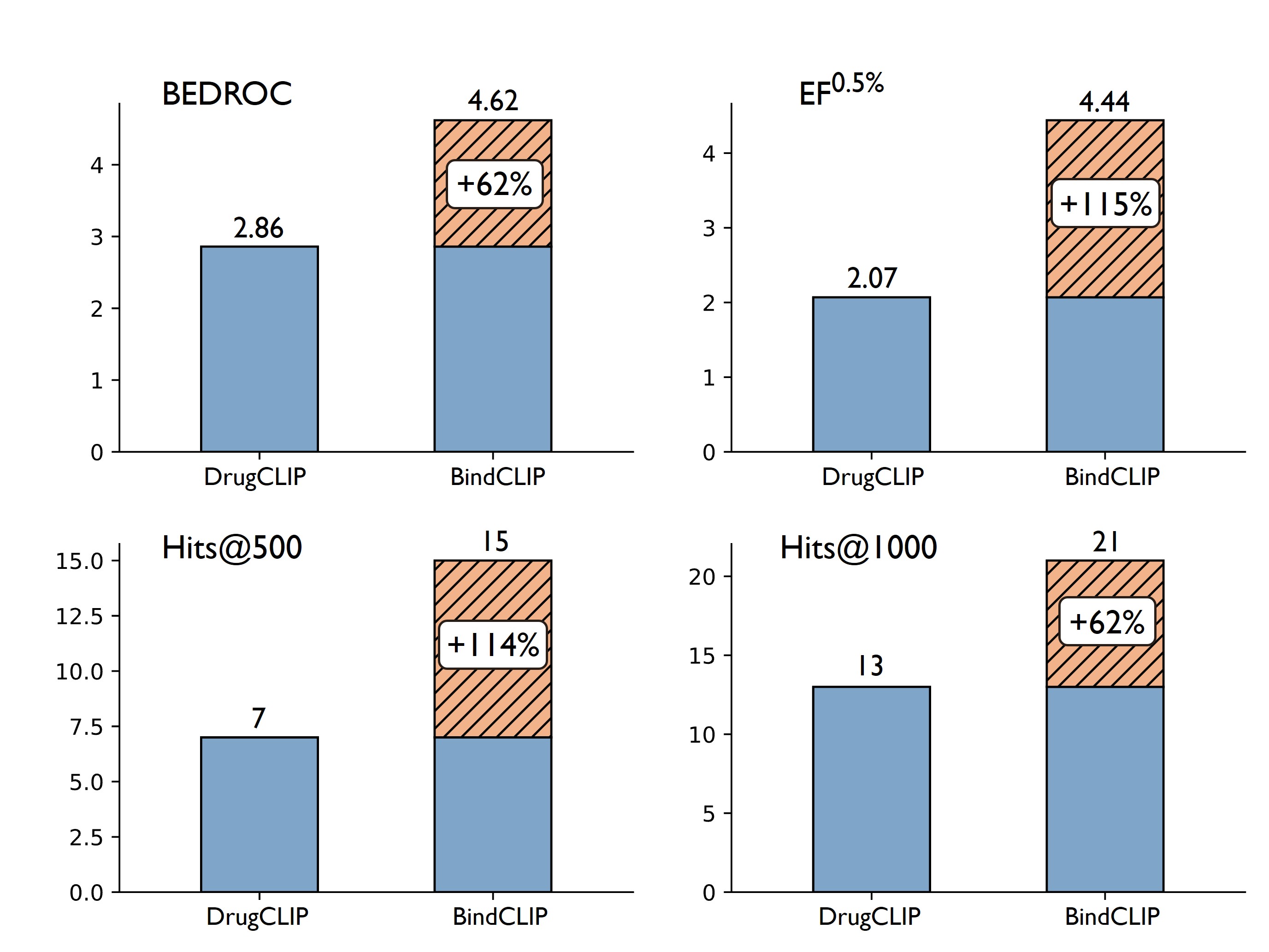}
    \caption{Out-of-distribution evaluation on the MF-PCBA subset, excluding assays whose proteins have more than 30\% sequence identity with any protein in the training set.}
    \vspace{-6pt}
    \label{fig:ood}
\end{figure}

\begin{figure}[t]
  \centering
  \includegraphics[width=\linewidth]{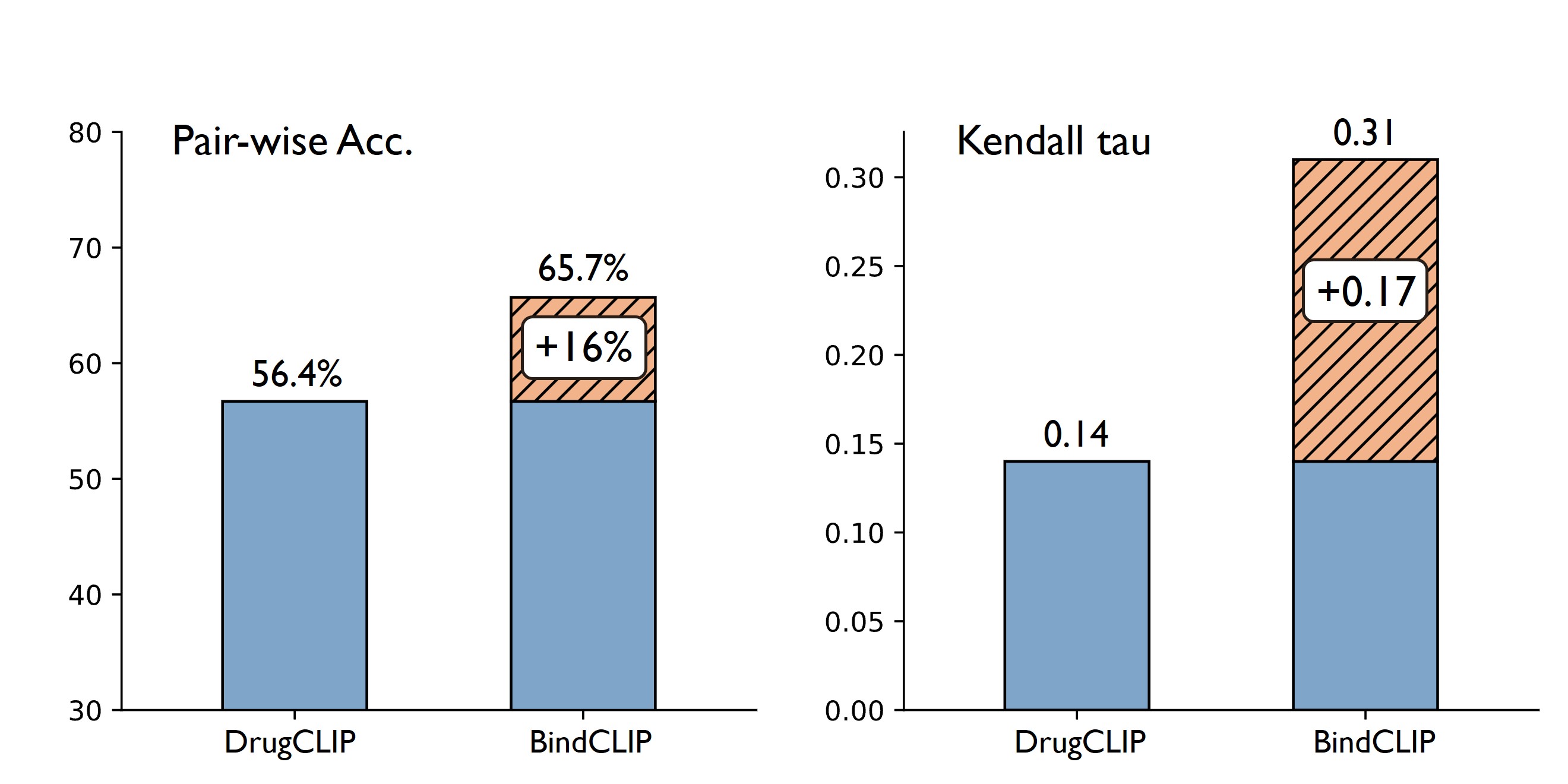}
    \caption{Evaluation on the 4-target FEP subset (CDK2, TYK2, JNK1, and P38) for ligand activity ranking within a chemical series.}
    \vspace{-8pt}
    \label{fig:fep}
\end{figure}

\noindent\textbf{Out-of-Distribution Generalization.}
We evaluate \ours on our constructed OOD benchmark, where actives account for only 0.7\% of the library, making retrieval particularly challenging in this imbalanced screening.
As shown in Fig.~\ref{fig:ood}, \ours consistently outperforms DrugCLIP, achieving substantial gains in early recognition,
with BEDROC improved by $+62\%$ and $\text{EF}^{0.5\%}$ by $+115\%$.
Moreover, at practical screening cutoffs, \ours retrieves significantly more active compounds:
Hits@500 improves by $+114\%$, and Hits@1000 by $+62\%$, highlighting its effectiveness in realistic screening scenarios.

\begin{table*}[t]
\caption{Ablation studies on LIT-PCBA.}
\label{table: ablation}
\begin{center}
\begin{tabular}{ccc|ccccc}
\toprule
\multicolumn{3}{c|}{Ablation Setting} 
& \multirow{2}{*}{AUROC}
& \multirow{2}{*}{BEDROC}
& \multirow{2}{*}{$\text{EF}^{0.5\%}$}
& \multirow{2}{*}{$\text{EF}^{1\%}$}
& \multirow{2}{*}{$\text{EF}^{5\%}$} \\
Diffusion & Rand. Neg. & Hard Neg. & & & & & \\
\midrule
  \xmark  & \xmark   &  \xmark  & 55.45$\pm$1.31 & 6.41$\pm$0.87 & 8.24$\pm$1.52 & 5.21$\pm$1.11 & 2.14$\pm$0.35 \\
 \cmark & \xmark   &  \xmark  & 56.71$\pm$1.10 & 6.85$\pm$0.41 & 9.11$\pm$0.67 & 5.72$\pm$0.32 & 2.27$\pm$0.22 \\
 \cmark   & \cmark &  \xmark  & 58.91$\pm$0.89 & 7.21$\pm$0.53 & 9.17$\pm$0.75 & 5.89$\pm$0.60 & 2.66$\pm$0.24 \\
\midrule
 \cmark & \xmark & \cmark & \textbf{59.15$\pm$1.26} 
& \textbf{7.88$\pm$0.25} 
& \textbf{9.84$\pm$0.65} 
& \textbf{6.26$\pm$0.35} 
& \textbf{2.90$\pm$0.18} \\
\bottomrule
\end{tabular}
\end{center}
\end{table*}

\noindent\textbf{Free Energy Perturbation--Based Ranking.}
FEP estimates the \emph{relative} binding free-energy difference between two ligands for a given target (i.e., $\Delta\Delta G$), and is widely regarded as a high-accuracy but challenging standard.
The 4-target FEP subset focuses on congeneric ligand series, which can be viewed as ranking a set of highly similar ligands that differ only by minor chemical modifications, making it particularly demanding to resolve subtle interaction changes and correctly order close analogues.

Fig.~\ref{fig:fep} reports performance on the FEP-based ranking task using pairwise accuracy and Kendall tau correlation.
\ours consistently outperforms DrugCLIP across both metrics, indicating improved fine-grained ranking capability among closely related ligands.
Specifically, \ours achieves a pairwise accuracy of 65.7\%, compared to 56.4\% for DrugCLIP, corresponding to a relative improvement of 16\%. This improvement is further reflected in Kendall tau correlation, which is computed from a global ranking induced by pairwise comparisons and increases from 0.14 to 0.31, indicating substantially improved consistency with the FEP-derived ordering.

We attribute these gains to the combination of diffusion supervision and anchored hard-negative contrastive learning, which enhance fine-grained interaction modeling
and thereby improve retrieval discriminability under distribution shift. Appendix Fig.~\ref{fig:activity_cliff} further shows an activity-cliff case.

\subsection{(RQ2) Ablation Study}
We conduct ablations on the more challenging LIT-PCBA benchmark to quantify the contribution of our two key training components: the pocket-conditioned diffusion objective and hard-negative augmentation. To better isolate the effect of hard negatives, we additionally include a random-negative variant, where decoy ligands are sampled uniformly from the molecular library.

Table~\ref{table: ablation} summarizes the results.
Starting from the baseline, adding the diffusion objective consistently improves performance across all metrics, with particularly notable gains in EF scores. This suggests that the binding pose-level diffusion provides an effective shaping signals that enrich contrastive training with interaction-relevant structural cues.
Adding randomly sampled negatives yields only marginal further improvements,
indicating that increasing negative diversity can be beneficial to some extent.
However, such random negatives are often trivially separable from true binders and therefore providing limited discriminative supervision compared to challenging negatives.
In contrast, augmenting training with hard negatives achieves the best overall performance.
 These results highlight the importance of informative negatives that are chemically similar to the positives, which encourages the model to learn more discriminative, interaction-aware representations.

\begin{table}[t]
\caption{
The performance of two probing experiments using linear probing on global (pocket--ligand affinity) and atom-level (atom--target distance) embeddings, averaged over five random seeds.
}
\vspace{-6pt}
\centering
\label{table:probe}
\begin{tabular}{cccc}
\toprule
Task                                                                                         & Model    & RMSE & MAE \\ \midrule
\multirow{2}{*}{\begin{tabular}[c]{@{}c@{}}Pocket--Ligand\\ Affinity Prediction\end{tabular}} & DrugCLIP  & 1.63   & 1.30    \\ \cline{2-4} 
                                                                                             & \ours & \textbf{1.35}   & \textbf{1.06}    \\ \midrule
\multirow{2}{*}{\begin{tabular}[c]{@{}c@{}}Atom--Target\\ Distance Prediction\end{tabular}}   & DrugCLIP  & 1.05   & 0.79    \\\cline{2-4} 
                                                                                             & \ours & \textbf{0.87}   & \textbf{0.61}    \\ \bottomrule
\end{tabular}
\centering
\end{table}

\subsection{(RQ3) Probing Fine-grained Interaction Knowledge.}
The pocket and ligand encoders produce both atom and global-level embeddings.
Although only the global embeddings are used for nearest-neighbor retrieval at inference time,
the atom-level features can still influence the quality of the global embeddings during training.
To examine whether our objectives inject finer-grained interaction knowledge into the learned embeddings,
we design two probing experiments:

(1) \textit{Atom--Pocket Distance Prediction} (atom-level). We train a linear probe that takes ligand atom embeddings together with a pooled embedding of pocket atoms, and predicts each ligand atom's minimum distance to the pocket.

(2) \textit{Pocket--Ligand Binding Affinity Prediction} (global-level). We train a linear probe over the global pocket and ligand embeddings to regress binding affinity.

As shown in Table~\ref{table:probe}, \ours yields consistent gains on both probing tasks,
suggesting that our training objectives encourage the embeddings to encode richer interaction signals.
On the global-level affinity regression probe, \ours achieves noticeably lower prediction error,
indicating that the learned global embeddings preserve richer interaction signals relevant to binding strength.
On the atom-level distance prediction probe, 
\ours more accurately recovers local geometric proximity between ligand atoms and the pocket.
Together, these results suggest that \ours injects fine-grained, spatially grounded binding cues into the representation space, improving both global interaction awareness and local geometry. See Appendix~\ref{appendix:probe} for details.

%% file: sections/appendix.tex
\section{Appendix}
\label{sec:appendix}

\subsection{Interaction-disrupting Operations.}
\label{appendix:motivation}
To probe whether a model's scoring truly depends on fine-grained binding cues rather than retrieval-level shortcuts, we construct \emph{interaction-disrupted} counterfactual ligands via minimal SMILES-level perturbations.
Given an original ligand SMILES $s$, we generate a modified variant $s^{\mathrm{disrupt}}$ that selectively disrupts common pocket--ligand interaction motifs while preserving overall physicochemical profiles.
Specifically, we implement four local \emph{interaction-disrupting operations} with a fixed priority from stronger to milder disruptions:
(i) \textbf{acid-to-amide} conversion $\mathrm{-C(=O)OH \rightarrow -C(=O)NH_2}$, which attenuates ionic and salt-bridge interactions;
(ii) \textbf{amine acetylation}, capping a non-amide, non-aromatic $\mathrm{N\!-\!H}$ into an acetamide $\mathrm{N\!-\!C(=O)CH_3}$;
(iii) \textbf{amine methylation}, converting $\mathrm{N\!-\!H}$ into $\mathrm{N\!-\!CH_3}$ to reduce hydrogen-bond donation;
and (iv) \textbf{hydroxyl methylation} $\mathrm{O\!-\!H \rightarrow O\!-\!CH_3}$ (excluding carboxylic acids), which masks hydroxyl hydrogen-bond donors.
For amine-based edits, we select a single eligible nitrogen atom that is non-aromatic, has at least one attached hydrogen, and is not amide-like, ensuring chemically valid substitutions.

To avoid confounding effects from large distributional shifts, we filter generated variants by physicochemical similarity between $s$ and $s^{\mathrm{disrupt}}$.
Concretely, we require
$|\Delta \mathrm{MW}|/\mathrm{MW}<\tau_{\mathrm{MW}}$,
$|\Delta \log P|<\tau_{\log P}$,
and $|\Delta \mathrm{tPSA}|<\tau_{\mathrm{tPSA}}$,
where $\mathrm{MW}$ denotes molecular weight, $\log P$ the octanol--water partition coefficient, and $\mathrm{tPSA}$ the topological polar surface area.
When multiple operations are applicable, we output at most one disrupted variant per ligand by taking the first valid transformation under the above priority order, and we canonicalize SMILES to deduplicate across edits.
This procedure yields paired $(s, s^{\mathrm{disrupt}})$ examples that preserve coarse physicochemical properties while selectively perturbing interaction-relevant functional groups, enabling controlled interaction-sensitivity analyses.

\subsection{Atom-level embedding visualization.}
\label{appendix:atomlevel}
To better understand the atom-level representations learned by CLIP-style virtual screening models, we visualize and compare ligand atom embeddings from DrugCLIP and UniMol~\cite{zhou2023_unimol}.
Since DrugCLIP is initialized from UniMol and further trained with a contrastive pocket--ligand objective, this comparison provides an intuitive view of how continued training influences atom-level embedding geometry.

\begin{figure}[t]
  \centering
  \includegraphics[width=\linewidth]{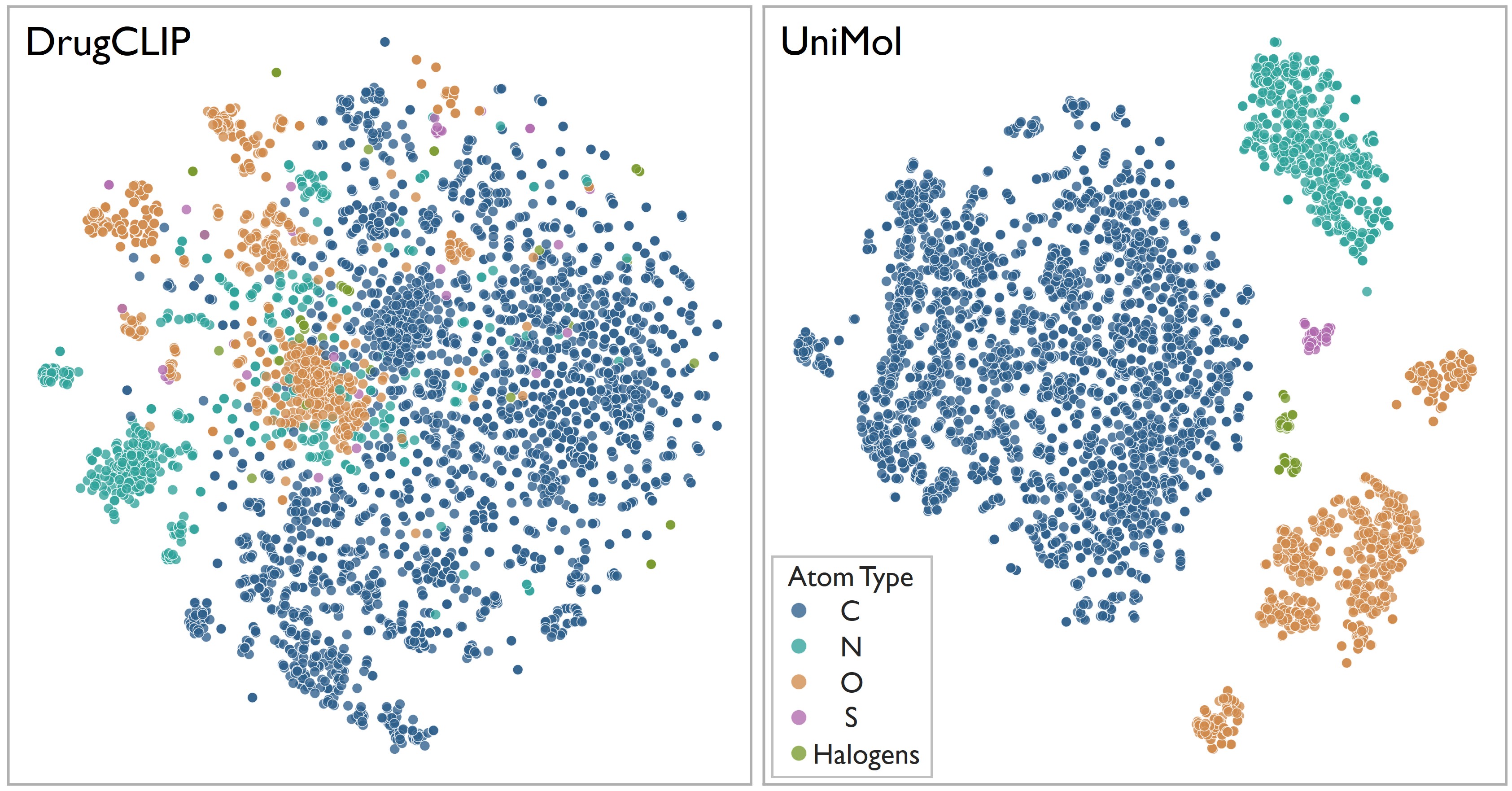}
    \caption{Atom-level embedding visualization (t-SNE). Ligand atom embeddings from DrugCLIP (left) and UniMol (right), colored by atom type (C/N/O/S/Halogens, where halogens include F, Cl, Br, and I).}
    \label{fig:atom}
\end{figure}

Specifically, we randomly sample 100 ligands from PDBbind 2019 and extract per-atom embeddings from both models.
We then visualize these atom embeddings via t-SNE, coloring each point by atom type.
In this visualization, UniMol embeddings exhibit clearer grouping patterns across atom types, whereas DrugCLIP embeddings appear more dispersed with less distinct separation.
This observation shows that continued contrastive training reshapes the geometry of atom-level representations.

\subsection{Vina-based ranking comparison.}
\label{appendix:vinaconsistency}
We examine model behavior at the global ranking level by comparing DrugCLIP's scoring-based ordering with a docking-based proxy (AutoDock Vina). We follow the official retrieval example released in the DrugCLIP GitHub repository, which provides a target pocket together with a ligand library for screening.

\begin{figure}[t]
  \centering
  \includegraphics[width=\linewidth]{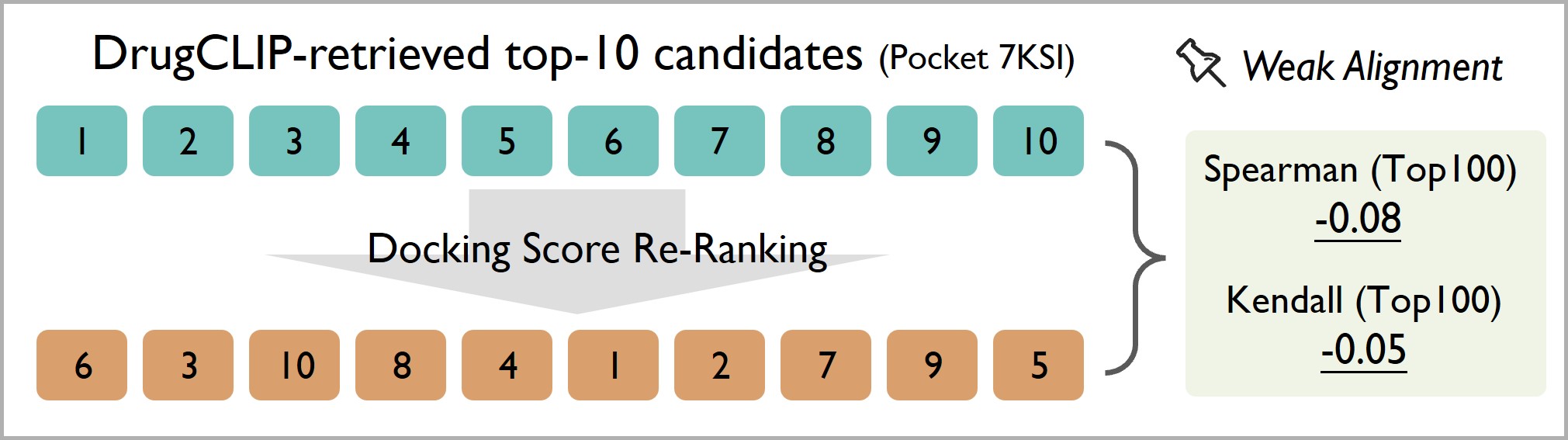}
    \caption{DrugCLIP ranking vs.\ Vina re-ranking. For pocket 7KSI, we take the top-100 candidates retrieved by DrugCLIP and re-rank them with AutoDock Vina (exhaustiveness=16).
}
    \label{fig:vina_rank}
\end{figure}

Using DrugCLIP, we retrieve the top-100 candidate ligands for the given pocket and obtain their predicted ranking scores. For the same candidates, we compute Vina scores (exhaustiveness=16) and re-rank the ligands accordingly.
As shown in Fig.~\ref{fig:vina_rank}, the Vina-induced ordering exhibits limited correspondence with the original DrugCLIP ranking.
Since Vina docking scores are an imperfect and approximate proxy, we include this comparison as a supplementary reference to illustrate potential differences between retrieval-based scoring and docking-based estimates.

This comparison also motivates our FEP+ ranking evaluation in the main paper. Since docking scores are only an approximate proxy, we further assess whether the induced rankings align with experimentally derived $\Delta\Delta G$ relations in the FEP+ benchmark~\cite{ross2023_fep+}, providing a complementary perspective on global ranking consistency.

\subsection{Activity-Cliff Case Study.}
\label{appendix:fepcase}
To qualitatively illustrate fine-grained ranking behavior, we visualize an activity-cliff example from the 4-target FEP subset. For JNK1, ligands \texttt{18634-1} and \texttt{18636-1} (subset identifiers) form a close analogue pair with a substantial free-energy gap ($\Delta\Delta G = 2.48$), indicating an activity cliff. As shown in Appendix Fig.~\ref{fig:activity_cliff}, DrugCLIP exhibits limited sensitivity to this subtle modification: the predicted scores for the two ligands are nearly indistinguishable, resulting in weak separation. In contrast, \ours responds more strongly and produces a noticeably larger score difference between \texttt{18634-1} and \texttt{18636-1}. This case suggests that diffusion supervision injects geometry-aware interaction cues into the embeddings, enabling \ours to better capture binding-relevant changes among close analogues.

\begin{figure}[h]
  \centering
  \includegraphics[width=\linewidth]{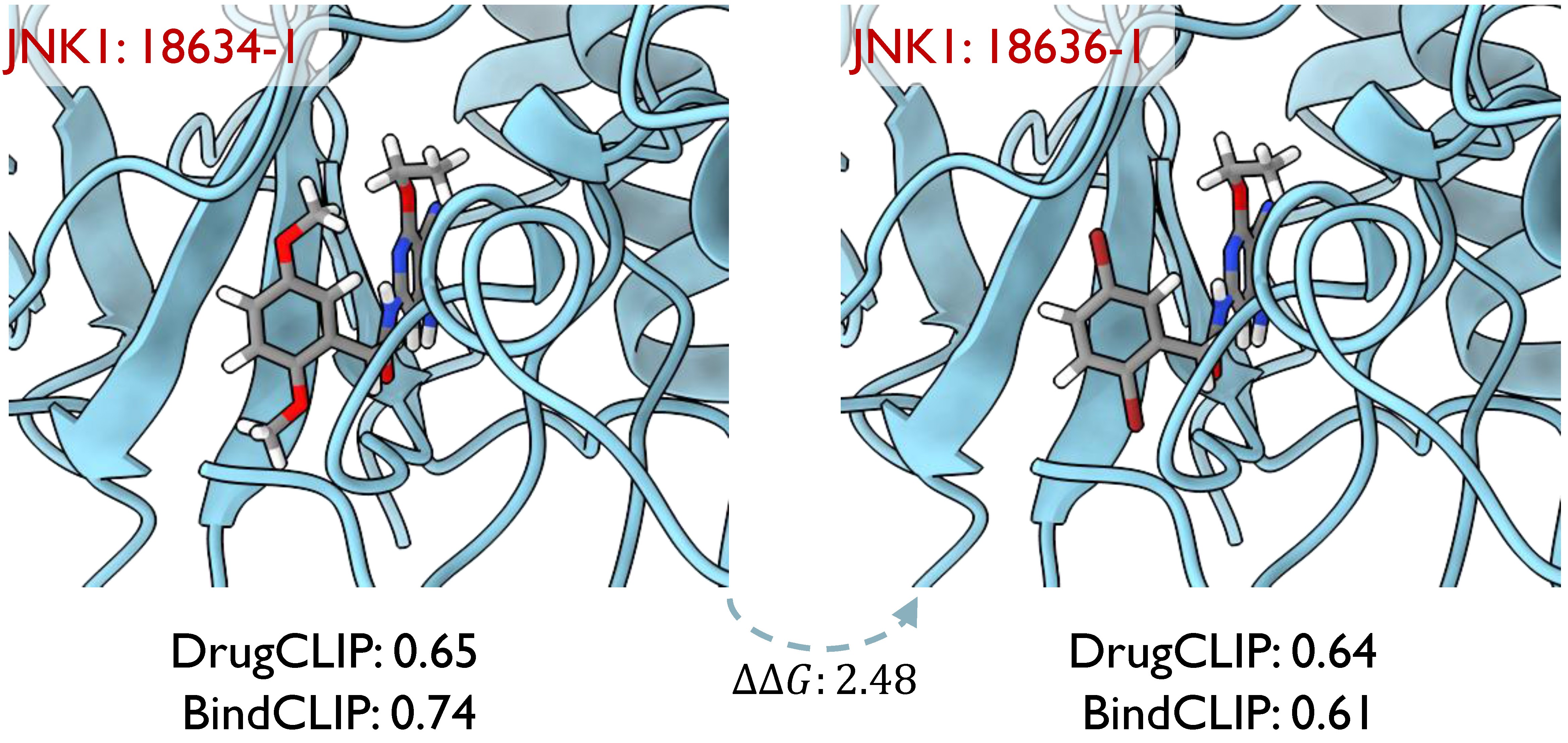}
    \caption{A case of an activity cliff in JNK1. Two structurally similar ligands (18634-1 and 18636-1) exhibit a large difference in binding free energy ($\Delta\Delta G$=2.48 kcal/mol). BindCLIP better captures the affinity ordering across this activity cliff.
}
    \label{fig:activity_cliff}
\end{figure}

\subsection{Probe Details}
\label{appendix:probe}

\paragraph{Pocket--Ligand Binding Affinity Prediction.}
We evaluate whether the learned \emph{global} pocket and ligand embeddings retain information predictive of binding strength via a linear-probing protocol.
For each protein--ligand complex, we extract the global ligand embedding $\mathbf{z}_m$ and the global pocket embedding $\mathbf{z}_p$ from the \emph{frozen} encoders, concatenate them, and train a linear regressor
\(
\hat{y}=\mathbf{w}^{\top}[\mathbf{z}_m;\mathbf{z}_p]+b,
\)
where $y$ is the ground-truth affinity label.
We construct the probe dataset from PDBbind 2019~\cite{wang2005_pdbbind} by taking the reported $pK$ value (i.e., $-\log K_d/K_i$ when available) as the regression target; complexes without a matched affinity entry are excluded.
We perform a random complex split into train/validation/test sets with an 8/1/1 ratio, optimize the linear regressor on the training set using mean squared error, and select the best checkpoint based on validation performance.
We then report RMSE and MAE on the held-out test set.
All results are averaged over five random seeds.

\paragraph{Atom--Pocket Distance Prediction.}
We evaluate whether the learned \emph{atom-level} embeddings encode local, spatially grounded binding cues by predicting geometric proximity between ligand atoms and pocket atoms.
Specifically, we use the (docked) ligand coordinates and pocket coordinates, remove hydrogen atoms from both ligand and pocket, and compute the Euclidean distance matrix between all ligand--pocket atom pairs.
We collect candidate pairs within a fixed cutoff (6~\AA) and uniformly sample 10 pairs per complex to avoid over-representing complexes with many atoms.
For each sampled pair $(i,j)$, we form the input feature by concatenating the corresponding ligand and pocket atom embeddings, $\mathbf{x}_{ij}=[\mathbf{h}_{m,i};\mathbf{h}_{p,j}]$, and use the ground-truth distance as the regression target.
We train a linear probe head on these fixed embedding, optimizing mean squared error.
We perform a random split of the atom-pair dataset into train/validation/test sets with an 8/1/1 ratio, select the best checkpoint based on validation RMSE, and report RMSE and MAE on the held-out test set.
All results are averaged over five random seeds.

\subsection{Diffusion Details}
\label{appendix:diffusion}

\noindent\textbf{Forward process.}
We define a diffusion process over the continuous ligand coordinates while treating the pocket as a fixed condition.
At diffusion step $t\in\{1,\ldots,T\}$, we corrupt $\mathbf{x}^{(t-1)}$ by adding Gaussian noise with a fixed variance schedule $\{\beta_t\}_{t=1}^{T}$~\cite{hoogeboom2021_scheme}:
\begin{equation}
q(\mathbf{x}^{(t)} \mid \mathbf{x}^{(t-1)})=
\mathcal{N}\!\left(\mathbf{x}^{(t)};\sqrt{1-\beta_t}\,\mathbf{x}^{(t-1)}, \beta_t \mathbf{I}\right),
\label{eq:forward_xt}
\end{equation}
where $\mathbf{I}$ denotes the identity matrix on $\mathbb{R}^{N_m\times 3}$ (i.e., isotropic noise applied independently to each coordinate dimension).

Let $\alpha_t = 1-\beta_t$ and $\bar{\alpha}_t=\prod_{s=1}^{t}\alpha_s$.
Then the marginal $q(\mathbf{x}^{(t)}\mid \mathbf{x}^{(0)})$ admits a closed form:
\begin{equation}
q(\mathbf{x}^{(t)}\mid \mathbf{x}^{(0)})=
\mathcal{N}\!\left(\mathbf{x}^{(t)};\sqrt{\bar{\alpha}_t}\,\mathbf{x}^{(0)}, (1-\bar{\alpha}_t)\mathbf{I}\right),
\label{eq:marginal_xt}
\end{equation}
equivalently,
\begin{equation}
\mathbf{x}^{(t)}=\sqrt{\bar{\alpha}_t}\,\mathbf{x}^{(0)}+\sqrt{1-\bar{\alpha}_t}\,\boldsymbol{\epsilon},
\qquad \boldsymbol{\epsilon}\sim\mathcal{N}(\mathbf{0},\mathbf{I}).
\label{eq:closedform_sample}
\end{equation}
In training, we sample $t\sim\mathcal{U}(\{1,\ldots,T\})$ and obtain $\mathbf{x}^{(t)}$ using Eq.~\ref{eq:closedform_sample}.

\noindent\textbf{Denoising network and timestep sampling.}
For the denoising model, we adopt the SE(3)-equivariant architecture of TargetDiff~\cite{guan_targetdiff} and set the total number of diffusion steps to $T=1000$.
While the training practice samples $t$ uniformly from $\{1,\ldots,T\}$ in DDPM-style models, we follow the representation-learning intuition highlighted in SODA~\cite{hudson2024_soda}, which suggests that intermediate noise levels tend to provide more informative learning signals than extremely low- or high-noise regimes.
Concretely, during training we sample
$t \sim \mathcal{U}(\{300, \ldots, 700\}),
\label{eq:t_sampling_mid}
$
and construct $\mathbf{x}^{(t)}$ via Eq.~\ref{eq:closedform_sample}.

\subsection{Training details.}
\label{appendix:train}
We train the retrieval model with a UniMol backbone in the Unicore framework.
Pocket and ligand encoders are initialized from official UniMol checkpoints.
All runs use a single NVIDIA A800 80G GPU.
We use Adam $(\beta_1,\beta_2)=(0.9,0.999)$ with $\epsilon=10^{-8}$ and gradient clipping (max norm 1.0), together with a polynomial learning-rate schedule (warmup ratio 0.06, base LR $10^{-3}$).
We train for 100 epochs with batch size 48, gradient accumulation of 4 steps, and validation batch size 64.
We select checkpoints by validation BEDROC and report mean$\pm$std over five seeds.

\subsection{Efficiency.}
\label{appendix:efficiency}
We train \ours on a single NVIDIA A800 80GB GPU with batch size 48. Compared to DrugCLIP, training \ours increases peak GPU memory usage by 36\% (1 hard negative) and 60\% (3 hard negatives), and takes 2.0$\times$ and 2.2$\times$ the wall-clock time, respectively. Importantly, inference remains identical to DrugCLIP: we only run the pocket and ligand encoders and perform nearest-neighbor retrieval in the embedding space, without any diffusion sampling or additional modules, and thus incur no additional inference overhead.